\documentclass[sigconf]{acmart}

\AtBeginDocument{%
  }

\acmConference[MMIR23@ACM-2023]{Workshop of 1st MMIR  Deep Multimodal Learning for Information Retrieval}{October 29 - November 3, 2023}{Ottawa, Ontario, Canada}

\usepackage{soul}
\usepackage{amsmath}
\usepackage{multirow}
\usepackage{tabularx}
\usepackage{graphicx}
\usepackage{stfloats}
\DeclareMathOperator{\Lagr}{\mathcal{L}}
\usepackage{subfig}
\usepackage{amsmath}

\begin{document}

\copyrightyear{2023}
\acmYear{2023}
\setcopyright{acmlicensed}\acmConference[MMIR '23]{Proceedings of the
1st International Workshop on Deep Multimodal Learning for Information
Retrieval}{November 2, 2023}{Ottawa, ON, Canada}
\acmBooktitle{Proceedings of the 1st International Workshop on Deep Multimodal Learning for Information Retrieval (MMIR '23), November 2, 2023, Ottawa, ON, Canada}
\acmPrice{15.00}
\acmDOI{10.1145/3606040.3617444}
\acmISBN{979-8-4007-0271-6/23/11}

\title{TC-OCR: TableCraft OCR for Efficient Detection \& Recognition of Table Structure \& Content}

\author{Avinash Anand}
\email{	avinasha@iiitd.ac.in}
\affiliation{%
  \institution{Indraprastha Institute of Information Technology, Delhi, MIDAS}
  \city{New Delhi}
  \country{India}}

\author{Raj Jaiswal}
\authornotemark[1]
\thanks{* These authors contributed equally to this work}
\email{Jaiswalofficial2908@gmail.com}
\affiliation{%
  \institution{Indraprastha Institute of Information Technology, Delhi, MIDAS}
  \city{New Delhi}
  \country{India}}

\author{Pijush Bhuyan}
\authornotemark[1]
\email{pijush22049@iiitd.ac.in}
\affiliation{%
  \institution{Indraprastha Institute of Information Technology, Delhi, MIDAS}
  \city{New Delhi}
  \country{India}}

\author{Mohit Gupta}
\authornotemark[1]
\email{mohit22112@iiitd.ac.in}
\affiliation{%
  \institution{Indraprastha Institute of Information Technology, Delhi, MIDAS}
  \city{New Delhi}
  \country{India}}

\author{Siddhesh Bangar}
\email{siddheshb008@gmail.com}
\affiliation{%
  \institution{Indraprastha Institute of Information Technology, Delhi, MIDAS}
  \city{New Delhi}
  \country{India}}

\author{Md. Modassir Imam}
\email{modimam123@gmail.com}
\affiliation{%
  \institution{Indraprastha Institute of Information Technology, Delhi, MIDAS}
  \city{New Delhi}
  \country{India}}

\author{Rajiv Ratn Shah}
\email{rajivratn@iiitd.ac.in}
\affiliation{%
  \institution{Indraprastha Institute of Information Technology, Delhi, MIDAS}
  \city{New Delhi}
  \country{India}}

\author{Shin'ichi Satoh}
\email{satoh@nii.ac.jp}
\affiliation{%
  \institution{National Institute of Informatics}
  \city{Tokyo}
  \country{Japan}}

\renewcommand{\shortauthors}{Avinash Anand et al.}

\begin{abstract}
The automatic recognition of tabular data in document images presents a significant challenge due to the diverse range of table styles and complex structures. Tables offer valuable content representation, enhancing the predictive capabilities of various systems such as search engines and Knowledge Graphs. Addressing the two main problems, namely table detection (TD) and table structure recognition (TSR), has traditionally been approached independently. In this research, we propose an end-to-end pipeline that integrates deep learning models, including DETR, CascadeTabNet, and PP OCR v2, to achieve comprehensive image-based table recognition. This integrated approach effectively handles diverse table styles, complex structures, and image distortions, resulting in improved accuracy and efficiency compared to existing methods like Table Transformers. Our system achieves simultaneous table detection (TD), table structure recognition (TSR), and table content recognition (TCR), preserving table structures and accurately extracting tabular data from document images. The integration of multiple models addresses the intricacies of table recognition, making our approach a promising solution for image-based table understanding, data extraction, and information retrieval applications. Our proposed approach achieves an IOU of 0.96 and an OCR Accuracy of 78\%, showcasing a remarkable improvement of approximately 25\% in the OCR Accuracy compared to the previous Table Transformer approach.
\end{abstract}



\maketitle

\section{Introduction}
As the global digital transformation continues to progress, there is a notable and accelerating trend toward replacing traditional physical paper-based documents with their digitized counterparts. These digital documents frequently contain tables that display various formats and layouts, presenting a diverse range of information. Tables play a pivotal role in succinctly conveying extensive data, allowing readers to efficiently explore, compare, and comprehend the content. Nevertheless, the compact nature of tables often presents significant challenges for machine parsing and comprehension processes.

Automatic Information Extraction from tables involves two essential sub-tasks Table Identification and Table Structure Recognition. Several studies \cite{schreiber2017deepdesrt, hao2016table, tran2015table, gilani2017table, traquair2019deep} have made significant contributions to the advancement of table detection, while others \cite{mao2003document, kara2020holistic, sarkar2020document} have focused on improving table structure recognition. These tasks are of utmost importance in the field of image analysis as they facilitate the extraction of critical information from tables in a digital format. Table detection is concerned with accurately identifying the precise spatial region within an image that contains the table. Conversely, table structure recognition involves the precise identification of table rows and columns, thereby enabling the extraction of individual table cells.

In the field of table recognition (TR), To efficiently use data from table images, computer vision-based pattern recognition methods are used. Table detection (TD), table structure recognition (TSR), and table content recognition (TCR) are the three main tasks involved in TR. TD focuses on locating tables within images, TSR aims to recognize their internal structures, and TCR involves extracting textual contents from the tables. The current emphasis is on developing end-to-end TR systems capable of seamlessly integrating all three sub-tasks. The primary goal is to address real-world scenarios where the system performs TD, TSR, and TCR simultaneously, thus enhancing the efficiency and effectiveness of table recognition in practical applications.

Despite the advancements in current open-source and commercial document analysis algorithms, such as the Table Transformer Model, certain limitations persist. For instance, due to the computational complexity and maximum sequence length constraint of Transformers, capturing long-range dependencies between cells can be challenging. As a result, lengthy tables may suffer from information loss, affecting the model's ability to understand the context accurately. Additionally, when encountering tables with numerous empty cells or sparse content, the model might struggle to distinguish meaningful empty cells from those with missing data. To address these limitations, we present our innovative solution that aims to overcome these challenges and enhance the overall performance of table analysis and recognition.

With the help of our proposed approach, table extraction methods can gain a better understanding of the inherent characteristics of tables, leading to improved accuracy in detecting and extracting table structures from document images. The main contributions of this paper can be summarized as follows:
\begin{itemize}
     \item We have proposed a novel integrated pipeline that combines three state-of-the-art models: DETR, CascadeTabNet, and PP OCR v2, to achieve end-to-end table recognition from image-based data. This innovative pipeline effectively addresses the significant challenges posed by variations in table styles, intricate structures, and image distortions commonly encountered in document images.
     \item Through rigorous experimentation and evaluation, we have demonstrated that our integrated pipeline outperforms existing methods in terms of both accuracy and efficiency for table recognition. The results highlight the pipeline's remarkable ability to preserve complex table structures and accurately extract tabular data from document images. These findings contribute to the advancement of image-based table recognition techniques and offer practical insights for handling diverse table layouts in real-world scenarios.
 \end{itemize}

\section{Related Work}
The task of table structure identification has been a challenging and unresolved issue within the document-parsing community, leading to the organization of several public challenges to address it \cite{dejeanicdar,gobel2013icdar,kayal2021icdar}. The difficulty of this problem can be attributed to various factors.

Firstly, tables exhibit a wide range of shapes and sizes, necessitating a flexible approach to effectively handle their diversity. This is particularly crucial when dealing with complex column and row headers, which can be highly intricate and demanding. Secondly, one of the complexities arises from the scarcity of data specifically tailored for table structure analysis. Nevertheless, there has been significant progress in recent years with the introduction of valuable datasets such as PubTabNet \cite{zhong2019image}, FinTabNet \cite{zheng2020global}, and TableBank \cite{li2019tablebank}, addressing this data deficiency.

\subsection{Table Detection}
Several significant contributions have been made in the field of table detection for document analysis. Hao et al. \cite{hao2016table} proposed a table detection method based on convolutional neural networks (CNN) specifically designed for PDF documents. Siddiqui et al. \cite{siddiqui2018decnt} introduced an innovative strategy that combines deformable CNN with faster region-based convolutional neural network (R-CNN) or feature pyramid network (FPN) to address the complexities arising from variable table sizes and orientations. Anand et al.~\cite{anand2024ranlaynet} proposes a noisy document images dataset for document layout detection, and shown improved performance in detecting tables in the document image.

Holevcek et al. \cite{holevcek2019table} extended the application of graph neural networks to structured documents, focusing on bills, where they utilized graph convolutions to facilitate table understanding. Casado et al. \cite{casado2020benefits} extensively explored object detection techniques, including Mask R-CNN, YOLO, SSD, and Retina Net, and demonstrated that fine-tuning from a domain closer to the target domain can significantly improve table detection performance.

Nguyen et al. \cite{nguyen2022tablesegnet} proposed TableSegNet, a compact fully convolutional network capable of simultaneously performing table separation and detection. Zhang et al. \cite{zhang2023yolo} introduced a YOLO-based table detection methodology, enhancing spatial arrangement learning through improving efficiency by including an involution into the network's core and using a straightforward feature pyramid network.

These studies collectively showcase the effectiveness of deep learning models, such as CNN and YOLO, in the context of table detection. Moreover, they highlight the benefits of incorporating specific techniques like deformable CNN, graph convolutions, and involution, which have proven instrumental in overcoming the inherent challenges associated with this task.

\subsection{Table Structure Recognition} 
Early approaches to table structure recognition heavily relied on hand-crafted features and heuristic rules \cite{itonori1993table, kieninger1998table, WANG20041479}. These methods were particularly suitable for simple table structures or predefined data formats. However, in recent times, inspired by the remarkable success of deep learning in various computer vision tasks like object detection and semantic segmentation, several novel deep learning-based methods \cite{raja2020table, schreiber2017deepdesrt} have emerged for table structure recognition.

Schreiber et al. (2017)~\cite{schreiber2017deepdesrt} introduced DeepDeSRT, a two-fold system that effectively combines Faster R-CNN and FCN for accurate table detection and precise row/column segmentation. On the other hand, Raja et al. (2020)~\cite{raja2020table} presented TabStruct-Net, an innovative Customized cell detection and interaction modules that precisely identify cells and anticipate their row and column relationships with other detected cells are incorporated into a framework for recognizing table structures.

These cutting-edge deep learning-based methods, exemplified by DeepDeSRT and TabStruct-Net, leverage the intrinsic capabilities of neural networks to significantly enhance table structure recognition by automatically learning relevant and discriminative features while capturing complex interrelationships within the tables.

\subsection{Table recognition}
Prior studies in table recognition have predominantly focused on non-end-to-end methodologies, dividing the problem into two distinct sub-tasks: table structure recognition and cell-content recognition. These approaches attempted to tackle each sub-problem independently using separate systems.

TableMASTER, introduced by \cite{ye2021pingan,He2021PingAnVCGroupsSF,lu2021master,li2018shape}, is a Transformer-based model specifically designed for table structure recognition. The method combines the Transformer model with a text line detector to identify text lines within individual table cells. Furthermore, they employed a text line recognizer based on the work of \cite{lu2021master} to extract text content from the identified lines.

Another Transformer-based model called TableFormer was proposed by \cite{nassartableformer}, which not only recognizes table structure but also predicts the bounding boxes of each table cell. These predicted bounding boxes were then utilized to extract the cell contents from PDF documents, resulting in a comprehensive table recognition system.

\begin{figure*}[ht]
\centering
\Description{Our TC-OCR achieves simultaneous Table Detection (TD), Table Structure Recognition (TSR), and Table Content Recognition (TCR), preserving table structures and accurately extracting tabular data from document images.}
  \includegraphics[width=0.9\linewidth]{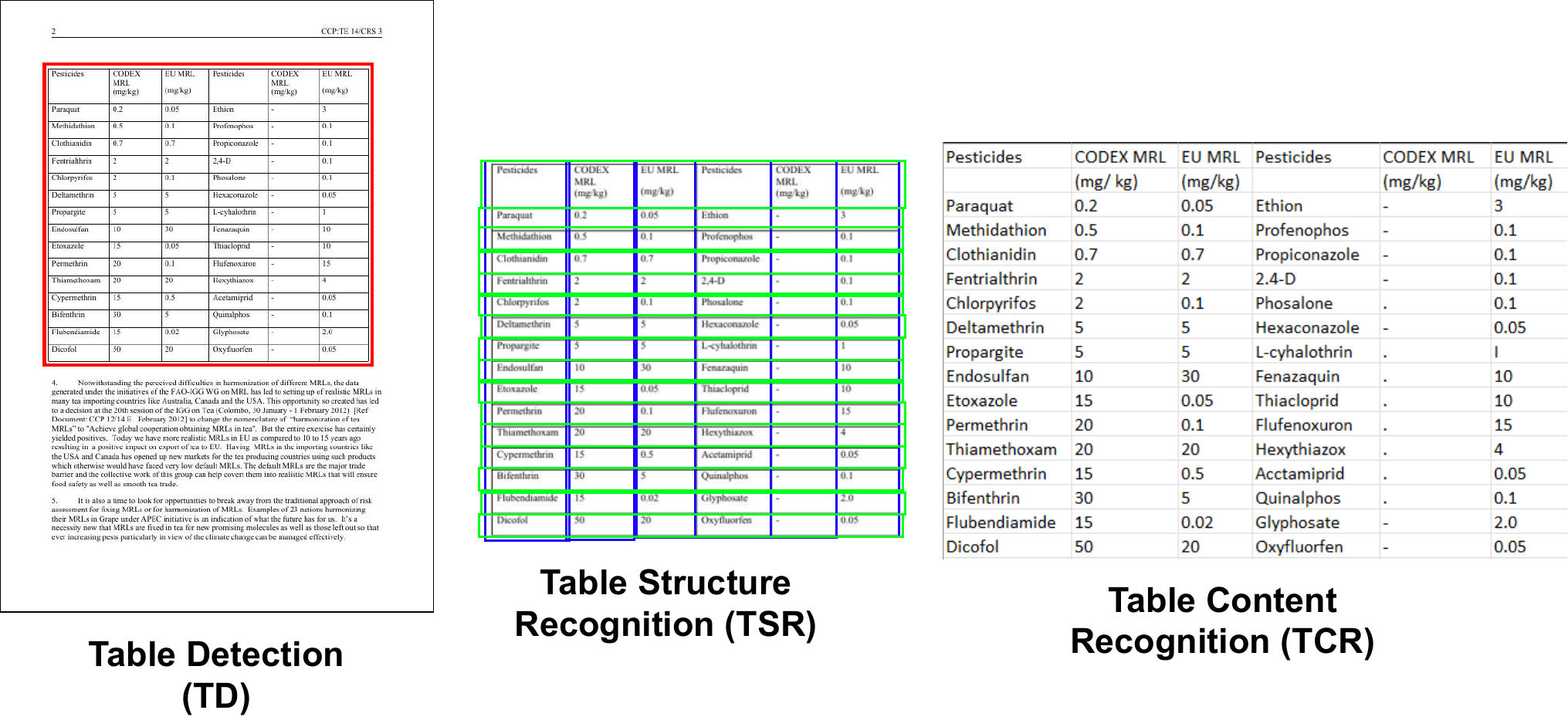}
  \caption{Our TC-OCR achieves simultaneous Table Detection (TD), Table Structure Recognition (TSR), and Table Content Recognition (TCR), preserving table structures and accurately extracting tabular data from document images.} 
\label{samples}
\end{figure*}

Recently, researchers have been shifting towards end-to-end approaches due to the advancements in deep learning and the increased availability of tabular data \cite{ly2023rethinking}. As an example, \cite{ly2023rethinking} introduced the encoder-dual decoder (EDD) model, which is capable of jointly recognizing both table structure and content for each cell. In addition to the model, they also introduced the PubTabNet dataset, which specifically focuses on table recognition and is made accessible to the research community.

Notably, the ICDAR2021 competition on scientific literature parsing organized by IBM Research in collaboration with IEEE ICDAR \cite{azeem2021current} has further contributed to advancements in table recognition.

In summary, the field of table recognition has witnessed significant progress through various techniques, from non-end-to-end to end-to-end approaches, and the development of new datasets and competitions has been instrumental in driving further advancements.

\section{Dataset}
Researchers have developed TableBank \cite{li2019tablebank}, an extensive standardized open-domain table benchmark dataset, to address the need for large-scale table analysis in various domains. The dataset surpasses existing human-labeled datasets in terms of size and contains 417,234 tables, each with its original document. TableBank includes a diverse range of domains, such as business documents, official filings, and research papers. The dataset is created by manipulating the mark-up tags for tables present in electronic documents like Microsoft Word (.docx) and LaTeX (.tex) files. Bounding boxes are added using the mark-up language to provide high-quality labeled data. The image-based table analysis approach used in TableBank is versatile, as it can handle different document types, including PDF, HTML, PowerPoint, and scanned versions. This robustness allows for the extraction of tables from various sources, enabling large-scale table analysis tasks.

\section{Methodology}
We have developed a comprehensive pipeline that integrates three distinct models to address various challenges associated with diverse table styles, complex structures, and image distortions commonly encountered in document images.

\begin{figure*}[ht]
\centering
\Description{Architecture of the proposed Methodology, where we have incorporated three distinct models DETR for table detection, CascadeTabNet for table structure recognition, and PP OCRv2 for text detection and recognition.}
  \includegraphics[width=\linewidth, height=5.5cm]{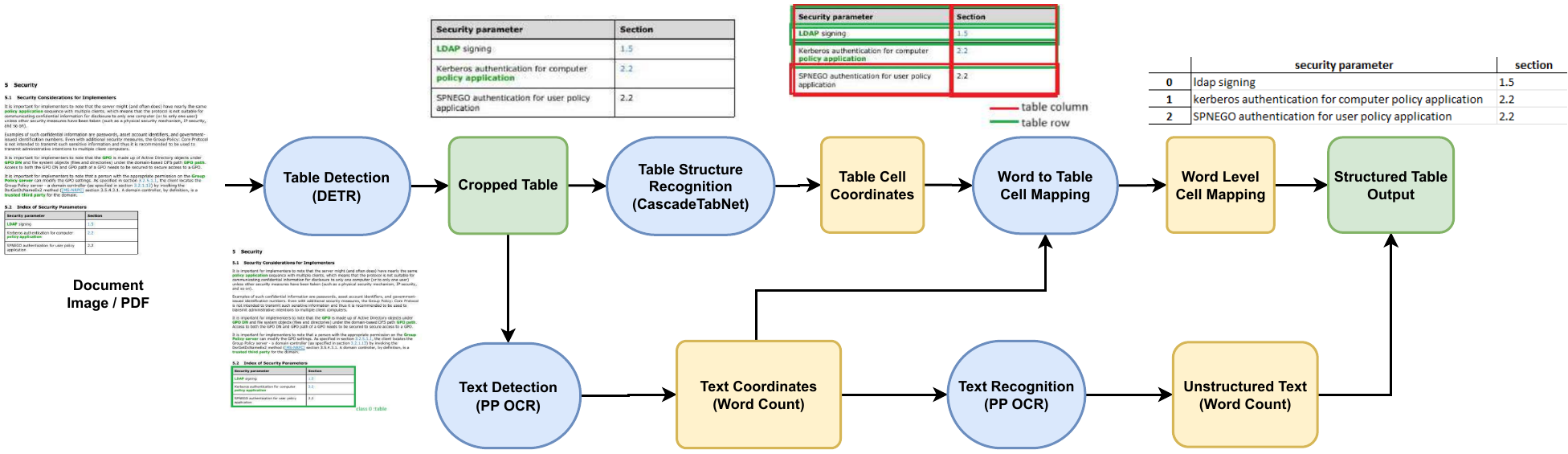}
\caption{Architecture of the proposed Methodology, where we have incorporated three distinct models DETR for table detection, CascadeTabNet for table structure recognition, and PP OCRv2 for text detection and recognition}
\label{model_pipeline}
\end{figure*}

\subsection{DETR - Object Detection Model}
\textbf{DE}tection \textbf{TR}ansformer or DETR \cite{carion2020end}, revolves around key elements, including a set-based global loss that ensures unique predictions through bipartite matching and a transformer encoder-decoder architecture. They presented a method for tackling object detection by formulating it as a direct set prediction problem. The approach employs an encoder-decoder architecture based on transformers, which are renowned for their effectiveness in sequence prediction tasks. Transformers \cite{vaswani2017attention} leverage self-attention mechanisms to explicitly model interactions between elements within a sequence. This characteristic makes transformers highly suitable for handling specific constraints in set prediction, such as eliminating duplicate predictions. 

By using this strategy, the detection pipeline reduced the requirement for manually created elements, such as anchor generation or non-maximum suppression processes, which frequently need prior task-specific expertise. We leverage this as an end-to-end, transformer-based solution for object detection, directly producing sets of bounding boxes and class labels. This ensures clear and distinct predictions, addressing issues related to duplicate detections. Moreover, the transformer encoder-decoder architecture significantly boosts detection performance by effectively capturing contextual relationships within the images.

\subsection{CascadeTabNet}
We used CascadeTabNet \cite{cascadetabnet2020}, an advanced end-to-end deep learning framework, which effectively tackles both table recognition sub-problems using a unified model. This methodology accomplishes pixel-level table segmentation, accurately identifying each table instance within an input image. Additionally, it performs table cell segmentation, predicting segmented regions corresponding to individual cells, thereby enabling the extraction of the table's structural information. The model accomplishes cell region predictions collectively in a single inference pass.

Moreover, the model has the capability to classify tables into two types: bordered (ruling-based) and borderless (no ruling-based) tables. For borderless tables, the model predicts cell segmentation directly.


The key components in the architecture involve leveraging the Cascade RCNN \cite{cai2018cascade}, which is a multi-stage model specifically designed to address the challenges of high-quality object detection in convolutional neural networks (CNNs). Additionally, a modified version of HRNet \cite{wang2019deep} is incorporated, providing reliable high-resolution representations and multi-level features that prove beneficial for the semantic segmentation tasks related to table recognition.

Through the fusion of these two approaches, CascadeTabNet achieves state-of-the-art performance in table recognition, effectively delivering precise table segmentation, cell segmentation, and accurate classification of table types.


\subsection{PP OCRv2}
The PP OCRv2 system~\cite{du2021ppocrv2} is designed to achieve high accuracy and computational efficiency for practical OCR applications. For text detection, the proposed Collaborative Mutual Learning (CML) and Copy Paste are two methods for data enhancement that have been successful in improving accuracy for object detection and instance segmentation tasks. CML involves training two student networks and a teacher network to develop a more robust text detector, and it also proves to be beneficial for text detection. Moreover, in-text recognition, they introduced the Lightweight CPU Network (PP-LCNet) \cite{cui2021pplcnet}, Unified-Deep Mutual Learning (U-DML), and CenterLoss. U-DML makes use of two student networks to improve text recognition precision. CenterLoss helps reduce errors caused by comparable characters.

We used the PP OCRv2 model to perform text-to-cell mapping in three phases. In the first phase, The mapping process links words to table cells $\mathcal{TC}$ using centroid coordinates $\mathcal{N}$, ensuring accurate associations within the table boundary. As shown in the equation below, where table cell centroid is denoted by $\mathcal{TCN}$.
\begin{equation}
    \mathcal{TCN}_{i,j} = \left(\frac{\mathcal{TC}_{i,j}(x_{1})+\mathcal{TC}_{i,j}(x_{2})}{2},\frac{\mathcal{TC}_{i,j}(y_{1}))+\mathcal{TC}_{i,j}(y_{2})}{2}\right)
\end{equation}
In the next phrase, A flexible threshold, set at half the cell’s width and length, accommodates variations in word positioning. Here it tries to calculate the centroid coordinates $\mathcal{ECN}$ for a text cell $k$ using the average of coordinates $\mathcal{EC}$ obtained from two reference points.
\begin{equation}
    \mathcal{ECN}_{k} = \left(\frac{\mathcal{EC}_{k}(x_{1})+\mathcal{EC}_{k}(x_{2})}{2},\frac{\mathcal{EC}_{k}(y_{1}))+\mathcal{EC}_{k}(y_{2})}{2}\right)
\end{equation}
Lastly, this approach preserves empty cells and avoids incorrect mappings, preventing text misalignment and enhancing word-to-cell precision.
\begin{equation}
    \mathcal{EC}_{k} = \left\{ \begin{array}{rcl} (\mathcal{R}_{i},\mathcal{C}_{j}), & if | \mathcal{EN}_{k}(x) - \mathcal{TN}_{i,j}(x)| \leq \frac{\mathcal{W}(\mathcal{T}_{i,j})}{2} \\
    & AND \\
    & | \mathcal{EN}_{k}(y) - \mathcal{TN}_{i,j}(y)| \leq \frac{\mathcal{W}(\mathcal{T}_{i,j})}{2} \\
    \phi, & \text{otherwise}
    \end{array}\right.
\end{equation}
\\
In the pipeline, we utilize PP OCRv2 for text detection and recognition purposes. The text cells detected by PP OCRv2 are compared with the cells identified by CascadeTabnet. Once a correspondence is found between the detected text and the cells, we calculate their centroids. By determining the minimum distance between any two cells, we are able to identify the structure or placement of the text within the rows $\mathcal{R}$ and columns $\mathcal{C}$ accurately.

In our proposed methodology for image-based table recognition, we present a comprehensive pipeline that incorporates three distinct models: DETR for table detection, CascadeTabNet for table structure recognition, and PPOCR for text detection and recognition as shown in figure \ref{model_pipeline}. This pipeline is specifically designed to tackle the challenges arising from various table styles, complex structures, and image distortions commonly encountered in document images.

Initially, the input document, which can be in image or PDF format, is preprocessed to ensure a standardized input for subsequent analysis. The document image is then fed into the DETR model, an object detection approach, which accurately localizes tables by generating a fixed-size set of S predictions. It is crucial for S to be bigger than the average number of things in the picture. During the loss computation, the model's training phase comprises an optimization procedure that creates the ideal bipartite matching between the predicted and ground truth items.

To address the limitations of existing table-structure identification models, we evaluated the Table Transformer \cite{smock2021tabletransformer}, which introduces a robust table-structure decomposition algorithm. This algorithm is designed to be language agnostic and effectively utilizes data from original PDF documents, enabling faster and more accurate text-cell extraction while establishing a direct link between table cells and their corresponding bounding boxes in the image. However, it is worth noting that the performance of the object detection decoder for table cells heavily relies on the availability of high-quality programmatic PDFs containing well-structured tabular content. In cases where the PDFs are poorly formatted or include non-standard table layouts, the model's performance may suffer, leading to less accurate content extraction.
\begin{equation}
    \hat{\sigma} = \arg\min_{\sigma \in T_S} \sum_{i=1}^{S} \Lagr_{\text{match}}(y_i, \hat{y}_{\sigma(i)})
\label{matchingcost}
\end{equation}

In Equation.~\ref{matchingcost}, $y$ denotes the ground truth set of objects, and $y = \{\hat{y}_i\}_{i=1}^S$ be the set of S predictions. $\Lagr_{match}(y_i, \hat{y}_{\sigma_(i)})$ is a pair-wise matching cost between ground truth $y_i$ and a prediction with index $\sigma(i)$.

The matching cost takes into account both the class predictions and the similarity of predicted and ground truth boxes. Each element $i$ of the ground truth set can be seen as $y_i = (c_i, b_i)$, where $c_i$ is the target class label and $b_i \in [0, 1]^4$ is a vector that specifies the height and width of the ground truth box in relation to the image's size as well as its center coordinates. For index $\sigma(i)$, we define probability of class $c_i$ as $\hat{p}_{\sigma(i)}(c_i)$, and predicted box as $\hat{b}_{\sigma(i)}$.
\begin{equation}
    \Lagr_{Hungarian}(y, \hat{y}) = \sum_{i=1}^{S}\left[ -log \hat{p}_{\hat{\sigma}(i)}(c_i) + 1_{(c_i\neq\emptyset)}\Lagr_{box}(b_i, \hat{b}_{\hat{\sigma}}(i))\right]
\label{hungarian}
\end{equation}
The next part of the matching cost and the Hungarian loss~\ref{hungarian} is $\Lagr_{box}(.)$ that scores the bounding boxes. In the Equation.~\ref{boxloss}, $l_1$ loss and the generalized IoU loss $\Lagr_{iou}$ is used, where $\lambda_{iou}, \lambda_{L1} \in \mathbb{R}$.
\begin{equation}
    \Lagr_{box}(b_i, \hat{b}_{\sigma(i)}) = \lambda_{iou}\Lagr_{iou}(b_i, \hat{b}_{\sigma(i)}) + \lambda_{L1} \| b_i - \hat{b}_{\sigma(i)} \|_1  
\label{boxloss}
\end{equation}

In this study, we propose a comprehensive approach for automatic table understanding in images. The process involves several key steps, starting with the detection of table regions through a region proposal technique. The identified table regions are then isolated from the original image and utilized as input for the CascadeTabNet model, a specialized deep-learning architecture designed for precise table structure recognition. CascadeTabNet is capable of accurately determining the number of rows and columns within a table and their corresponding spatial coordinates.

Subsequently, we employ the PPOCR (Pixel-level Patch-wise Object Character Recognition) method for precise text detection and recognition within the identified table cells. PPOCR extracts the spatial coordinates of the detected text, and we establish a mapping process based on the nearest neighbor approach to align this text with the original coordinates of the table cells obtained from CascadeTabNet.

This integrated methodology offers a robust and efficient solution for the automatic extraction and understanding of tabular data from images, enhancing the organization and accessibility of such information in various applications.
\begin{table*}[ht]
    \centering
    \renewcommand{\arraystretch}{1.5}
    \begin{tabularx}{0.95\textwidth}{p{1.8cm}|p{2cm}|p{1cm}|p{1cm}|p{1.3cm}|p{1.35cm}|p{1.5cm}|p{2cm}}
     \hline
     \hline
     \textbf{Column No.} & \textbf{No. of Images} & \textbf{Rows} & \textbf{TATR} & \textbf{TC-OCR} & \textbf{TATR Accuracy (\%)} & \textbf{TC-OCR Accuracy (\%)} & \textbf{Improvement (TC-OCR - TATR)} \\
     \hline
     \hline
     Total & 240 & 2785 & 1818 & \textbf{2485} & 65 & \textbf{89} & 24 \\
     \hline
     2 & 100 & 1130 & 838 & \textbf{1075} & 74 & \textbf{95} & 21 \\
     \hline
     3 & 100 & 1085 & 760 & \textbf{1010} & 70 & \textbf{91} & 21 \\
     \hline
     4 & 40 & 570 & 220 & \textbf{400} & 39 & \textbf{70} & 31 \\
     \hline
     \hline
     \end{tabularx}
     \vspace{2mm}
     \caption{Comprehensive comparison of results between the Table Transformer (TATR) model and our proposed method}\label{results}
\end{table*}
\begin{equation}
    Loss_{total} = Loss_{truth} + Loss_{dml} + Loss_{distill}
\label{totaldistill}
\end{equation}

Equation.~\ref{totaldistill} consists of three losses, 1) Truth Loss, 2) DML Loss, and 3) Distill Loss. Truth Loss is used to make sure that the training is supervised by the true label. Secondly, for calculating the DML Loss, KL Divergence is used which computes the distance between student models. Lastly, the third component is Distill Loss which reflects the supervision of the teacher model on the sub-student models.
\begin{equation}
    Loss_{total} = Loss_{ctc} + Loss_{dml} + Loss_{feat}
\label{totalloss}
\end{equation}

The total loss function~\ref{totalloss} consists of three parts~\cite{du2021ppocrv2} section 2.2:
\begin{itemize}
    \item \textbf{CTC Loss: } Due to the fact that both networks were trained from the beginning, they can converge using CTC loss.
    \item \textbf{DML Loss: } DML loss is necessary to make sure that the distributions of the two networks are consistent because it is predicted that the ultimate output distributions of the two networks would be identical.
    \item \textbf{Feature Loss: } It is expected that the two feature maps will be similar because the two network designs are similar. The gap between the intermediate feature maps of the two networks can be reduced via feature loss.
\end{itemize}

By leveraging the known structural characteristics of tables, we have devised a systematic pipeline for precise extraction of text in a structured manner from document images, while preserving the original table organization. The pipeline consists of three interconnected models: table localization, structure recognition, and structured text detection and recognition. The extracted data is then presented in a CSV file format, adhering to the same structure as the original table in the document. To compute the word-level accuracy $W_{Acc}$, the following formula is utilized.
\begin{equation}
    W_{Acc} = (\frac{X}{Y}) * 100
\label{word_level_accuracy}
\end{equation}
Where $X$ is the number of words correctly recognized by OCR, and $Y$ is the total number of words in the ground truth.


The proposed end-to-end solution demonstrates its effectiveness in image-based table recognition, addressing various challenges in the process. These challenges encompass table localization, structure recognition, and the accurate detection and recognition of text within the structured table. The successful implementation of this comprehensive approach allows for the accurate extraction of tabular data from document images, which in turn enhances data analysis, and search engine capabilities, and contributes to knowledge graph enrichment.

\section{Experiment}
We conducted a comparative analysis of inference time for our proposed model and Table Transformer (TATR) \cite{smock2021tabletransformer} on TableBank dataset \cite{li2019tablebank} comprising 47,053 table images as shown in table \ref{tab:inferencetime}. The table above presents the results of this evaluation. As observed, our model outperforms TATR in terms of efficiency, demonstrating faster inference times across all measured aspects. Specifically, our model achieves a maximum inference time of 12.7 seconds, a minimum of 5.42 seconds, and an average of 8.23 seconds. In contrast, TATR's corresponding figures are 15.48 seconds, 4.95 seconds, and 12.43 seconds, respectively. 

\setlength\tabcolsep{25pt}
\begin{table}[ht]
    \centering
    \renewcommand{\arraystretch}{1}
    \begin{tabular}{c|c|c}
    \hline
    \hline
         \textbf{Model} & \multicolumn{2}{c}{\textbf{Inference Time (sec.)}} \\
         \toprule
         \hline
         \multirow{3}{*}{TATR \cite{smock2021tabletransformer}} & Max & 15.48 \\
         & Min & 4.95 \\
         & Avg & 12.43 \\
         \midrule
         \multirow{3}{*}{TC-OCR} & Max & \textbf{12.7} \\   
         & Min & \textbf{5.42} \\
         & Avg & \textbf{8.23} \\
         \hline
         \bottomrule
    \end{tabular}
    \vspace{2mm}
    \caption{Inference time in seconds for Our model is compared against Table Transformer (TATR) on TableBank \cite{li2019tablebank} dataset of 47,053 images}
    \label{tab:inferencetime}
\end{table}
These findings underscore the effectiveness of our approach in jointly representing and integrating textual and visual information within tables, leading to enhanced performance and reduced inference times. The superior inference speed of our model positions it as a promising solution for real-world applications, where time-sensitive tasks demand swift and accurate data comprehension.

We also carried out a comparative analysis of our proposed model against the state-of-the-art (SOTA) Table Transformer model, we took 8,000 samples from TableBank \cite{li2019tablebank}. The table summarizes the evaluation results in terms of Intersection over Union (IOU) and Optical Character Recognition (OCR) accuracy.

As shown in Table.~\ref{tab:SOTAresults}, our model outperforms the Table Transformer in both metrics, showcasing its superior performance. Specifically, our model achieves an impressive IOU of 0.96, indicating its effectiveness in accurately delineating and localizing table elements. Moreover, our model demonstrates a significant advancement in OCR accuracy, reaching an impressive 78\%, thereby excelling in the crucial task of accurately recognizing and understanding the textual content within tables.

Another comprehensive comparison between the Table Transformer (TATR) model and our proposed method shown in table \ref{results} showcases the performance evaluation on different columns of a dataset containing a total of 240 images and 2,785 rows. Our method demonstrates superior accuracy across all columns, outperforming TATR significantly. Particularly noteworthy is the overall improvement achieved by our approach, with an impressive 24\% increase in accuracy compared to TATR. These findings underscore the effectiveness of our proposed method in tackling the problem of the multimodal table, indicating its potential for enhancing data comprehension and extraction of meaningful insights from diverse tabular data.


\section{Conclusion}

In conclusion, we propose an integrated pipeline for end-to-end image-based table recognition, leveraging the capabilities of three state-of-the-art models: DETR, CascadeTabNet, and PP OCR v2. By combining these models, we effectively tackle the challenges posed by diverse table styles and complex structures in document images. Our approach facilitates the accurate reconstruction of table layouts and the extraction of cell content from PDF or OCR through bounding boxes. Empirical evaluations demonstrate the superior performance and efficiency of our method compared to existing techniques, as it excels in preserving table structures and extracting tabular data with high efficacy. It is important to note that while our research serves as a strong foundation for advancing image-based table recognition, further refinements and optimizations are essential to enhance its applicability across a wider range of scenarios. Ultimately, our work contributes to the advancement of data extraction and comprehension in digitized documents, fostering innovation in the field of document analysis.

\setlength\tabcolsep{2pt}
\begin{table}[ht]
    \centering
    \setlength{\tabcolsep}{5\tabcolsep}
    \renewcommand{\arraystretch}{1.1}
    \begin{tabular}{l|c|c}
     \hline
     \hline
     \textbf{Model} & \textbf{IOU} & \textbf{OCR Accuracy} \\
     \toprule
     \hline
     \textbf{Table Transformer \cite{smock2021tabletransformer}} & 0.94 &  62 \% \\
     \textbf{Our Model (TC-OCR) } & \textbf{0.96} & \textbf{78 \%} \\
     \hline
     \bottomrule
     \end{tabular}
     \vspace{2mm}
     \caption{Comparison of our model (TC-OCR) with SOTA trained on 8000 samples of TableBank \cite{li2019tablebank} Dataset}
     \label{tab:SOTAresults}
\end{table}
\section{Future Scope}
The multi-modal tables problem presents a significant challenge in the realm of AI research, necessitating effective understanding and processing of tables that incorporate both textual and visual elements, such as images or graphs \cite{kostic2021multi}. Successfully addressing this challenge requires AI models to not only interpret the content within individual cells but also grasp the intricate relationships between textual and visual information.

Therefore, the primary objective of this research is to devise novel methods that can jointly represent and seamlessly integrate these modalities, leading to more comprehensive data comprehension and extraction of meaningful insights across diverse domains. By delving into this unexplored territory, this study aims to pave the way for innovative approaches that advance the capabilities of AI systems in handling multimodal tables and offer valuable contributions to real-world applications.

\section{ACKNOWLEDGMENT}
Dr. Rajiv Ratn Shah is partly supported by the Infosys Center for AI, the Center of Design and New Media, and the Center of Excellence in Healthcare at Indraprastha Institute of Information Technology, Delhi.
We sincerely appreciate the guidance and unwavering support provided by Ms. Astha Verma and Mr. Naman Lal throughout our research. Their expertise and insightful feedback have greatly influenced the direction and quality of our study. We are grateful for their time, dedication, and willingness to share knowledge, which significantly contributed to the completion of this work. Their encouragement and constructive discussions served as a constant source of motivation, and we feel privileged to have benefited from their wisdom and mentorship. 

\section{Limitations}
One notable limitation of our proposed approach is its inability to accurately recognize complex tables with merged cells, nested tables, or irregular structures. Dealing with such intricate layouts poses challenges in comprehending the intricate relationships between cells and headers. As a result, our current method may not be suitable for handling these specialized cases, and further research and enhancements are required to address these complexities effectively.

\bibliographystyle{ACM-Reference-Format}
\bibliography{software}
\end{document}